%% file: main.tex
\pdfoutput=1

\documentclass[11pt]{article}

\usepackage{EMNLP2022}

\usepackage{times}
\usepackage{latexsym}
\usepackage{fontawesome5}

\usepackage[T1]{fontenc}
\usepackage[russian,english]{babel}

\usepackage[utf8]{inputenc}

\usepackage{microtype}

\usepackage{inconsolata}

\usepackage{amssymb}
\usepackage{graphicx}
\usepackage{blindtext}
\usepackage{setspace}
\usepackage{multirow}
\usepackage{multicol}
\usepackage{float}
\usepackage{textcomp}
\usepackage{xcolor}
\usepackage{colortbl}
\usepackage{latexsym}
\usepackage{enumitem}
\usepackage{subfig}
\usepackage{wrapfig}
\usepackage{comment}
\usepackage{booktabs}
\usepackage{makecell}
\usepackage{tablefootnote}
\usepackage{kantlipsum}
\usepackage{soul}
\usepackage{enumitem}
\usepackage{longtable}
\usepackage[tableposition=below]{caption}
\usepackage[space]{grffile}

\definecolor{maroon}{RGB}{204,0,0}
\definecolor{green1}{RGB}{56,118,29}

\definecolor{light-gray}{gray}{0.95}
\newcommand{\code}[1]{\colorbox{light-gray}{\texttt{#1}}}

\newcommand{\myurl}[1]{\href{https://#1}{\UrlFont{#1}}}

\title{CLSE: Corpus of Linguistically Significant Entities}

\author{Aleksandr Chuklin\textsuperscript{*\ \faGoogle}, Justin Zhao\textsuperscript{*\ \faCubes}, Mihir Kale\textsuperscript{\ \faGoogle}\\
  \textsuperscript{\faGoogle}\ Google, \textsuperscript{\faCubes}\ Predibase \\
  \texttt{\{chuklin,mihirkale\}@google.com}, \texttt{justin@predibase.com}}

\begin{document}
\maketitle
\begin{abstract}
One of the biggest challenges of natural language generation~(NLG) is the proper handling of named entities.
Named entities are a common source of grammar mistakes such as wrong prepositions, wrong article handling,
or incorrect entity inflection.
Without factoring linguistic representation, such errors are often underrepresented when evaluating on a small set of arbitrarily picked argument values,
or when translating a dataset from a linguistically simpler language, like English, to a linguistically complex language, like Russian.
However, for some applications, broadly precise grammatical correctness is critical---native speakers may find entity-related grammar errors silly, jarring, or even offensive.

To enable the creation of more linguistically diverse NLG datasets,
we release a Corpus of Linguistically Significant Entities~(CLSE)
annotated by linguist experts. The corpus includes 34 languages and covers
74 different semantic types to support various applications from airline ticketing to video games.
To demonstrate one possible use of CLSE, we produce an augmented version of
the Schema-Guided Dialog Dataset, SGD-CLSE. Using the CLSE's entities
and a small number of human translations, we create a linguistically representative NLG evaluation benchmark
in three languages: French (high-resource), Marathi (low-resource), and Russian (highly inflected language).
We establish quality baselines for neural, template-based, and hybrid NLG systems
and discuss the strengths and weaknesses of each approach.
\end{abstract}

\def\thefootnote{*}\footnotetext{The first two authors contributed equally to this work.}
\def\thefootnote{\faCubes}\footnotetext{Part of this work done while at Google.}
\def\thefootnote{\arabic{footnote}}

\section{Introduction}

Natural language generation~(NLG)~\cite{reiter2000building} is an umbrella term for the problem of generating
fluent, human-like text from a variety of inputs. It covers, among others, text-to-text problems such as
text summarization or machine translation~\cite{allahyari2017text,stahlberg2020neural,malmi2022text},
image-to-text problems~\cite{hossain2019comprehensive}, and structured data to text~\cite{kukich1983design,mckeown1985text}.

\begin{figure}[tb]
\centering
\includegraphics[width=1.0\linewidth]{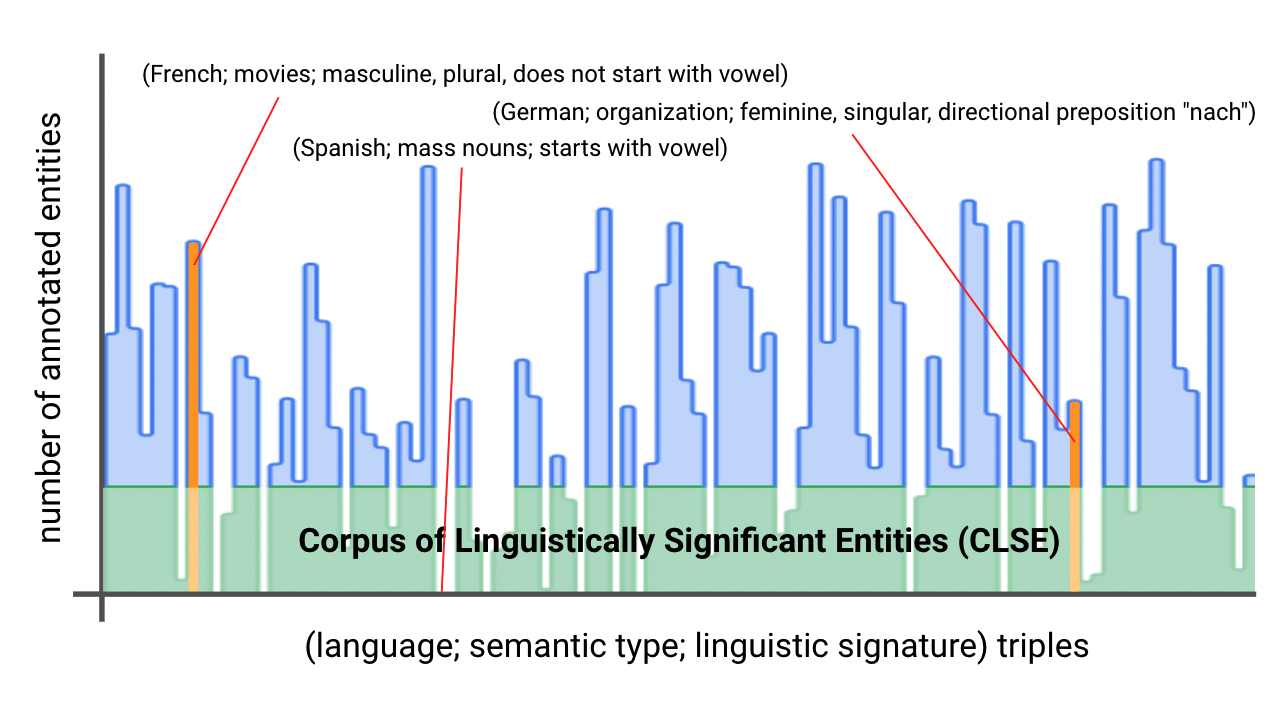}
\caption{The corpus of linguistically significant entities~(CLSE) is created by annotating a large number of entities with their linguistic properties. Entities are grouped by ($language$, $semantic\_type$, $linguistic\_signature$) triples. This results in a corpus of entities that, for a given language and semantic type, is balanced across linguistic properties. Note that some linguistic signatures have few or no annotated entities because they simply do not occur.%
}
\label{fig:clse_illustration}
\end{figure}

Unlike most natural language processing tasks, it is hard to produce
reusable ground truth labeled data for NLG:
there can be many different outputs that are grammatical and human-like.
This is why it is important to invest in NLG resources, especially for non-English languages.
The GEM benchmark~\cite{gehrmann2021gem} is such an initiative for NLG evaluation. Its second version
\cite{gehrmann2022gemv2} covers 40 tasks and 51 languages and provides an easy way for others to contribute.

In parallel to adding new NLG resources, a framework for dataset augmentation has recently been introduced
by~\citet{dhole2021nl}. The main idea is to perform transformation and slicing of the existing datasets
to boost certain properties as a way to adversarially test the generalization abilities of the benchmarked models.
Our contribution is similar in spirit: the corpus of linguistically significant entities~(CLSE) enables transforming existing datasets to be more balanced by linguistic phenomena covered by named entities (Figure~\ref{fig:clse_illustration}).

Ensuring grammatical correctness is arguably a basic requirement for any NLG system
in any language. Native speakers may find grammatical errors produced by such systems silly, jarring, or even offensive, for example if the utterance refers to an entity with the wrong tense, formality, animacy, or gender~\cite{dinan2020queens}.
The CLSE resource we introduce here is particularly useful for stress-testing the linguistic robustness of NLG systems.

The rest of the paper is organized as follows. Section~\ref{sec:clse} introduces the CLSE corpus annotated by linguist experts.
In Section~\ref{sec:sgd}, we discuss the problem of NLG for task-oriented dialogs and how CLSE can be applied to it.
In Section~\ref{sec:sgd_clse} we show how to construct a linguistically diverse dataset for this task, a dataset we refer to as SGD-CLSE.
In Section~\ref{sec:experiment}, we introduce three different baseline NLG systems
which we evaluate on this dataset. Results are discussed in Section~\ref{sec:results}.

\begin{table}[tb]
\begin{tabular}{l@{~}l@{~}l}
\textbf{case}              & \textbf{singular}                         & ~\textbf{plural}       \\
nominative    & \foreignlanguage{russian}{книга} \small{(kniga)}                    & \foreignlanguage{russian}{ книги} \small{(knigi)}     \\
genitive      & \foreignlanguage{russian}{книги} \small{(knigi)}                    &\foreignlanguage{russian}{ книг} \small{(knig)}       \\
dative        & \foreignlanguage{russian}{книге} \small{(knige)}                    &\foreignlanguage{russian}{ книгам} \small{(knigam)}   \\
accusative    & \foreignlanguage{russian}{книгу} \small{(knigu)}                    &\foreignlanguage{russian}{ книги} \small{(knigi)}     \\
instrumental  & \foreignlanguage{russian}{книгой} \small{(knigoy)}
&\foreignlanguage{russian}{ книгами} \small{(knigami)} \\
prepositional & \foreignlanguage{russian}{книге} \small{(knige)}                    &\foreignlanguage{russian}{ книгах} \small{(knigakh)} \\
\end{tabular}
\caption{Inflections of the word "book" in Russian. Cf. \foreignlanguage{russian}{"Я купил <книгу>"} ("I bought a <book>"; dative case) and
\foreignlanguage{russian}{"Моя <книга> потерялась"} ("My <book> got lost; nominative case).}
\label{tab:morph_inflection}
\end{table}

\section{Corpus of Linguistically Significant Entities (CLSE)}
\label{sec:clse}
The Google Knowledge Graph API\footnote{\myurl{developers.google.com/knowledge-graph}}~\cite{guha2016schema}
provides access to millions of entries that describe real-world entities like people, places, and things.
Each entity is a node in the graph and can be associated with any number of \href{https://schema.org}{schema.org} semantic types,
such as Person, AdministrativeArea, or TouristAttraction.

We first source lexical annotations from expert linguists for a large number of entities in the knowledge graph. Lexical annotations are language-specific and pertain to broader categories of linguistic properties like
Animacy,\footnote{\myurl{en.wikipedia.org/wiki/Animacy}}
Case,\footnote{See Table~\ref{tab:morph_inflection} as an illustration.}
Classifier, Countability, Definiteness, Gender, and Number.
Different languages consists of different linguistic properties.
For example, the concept of animacy is not used in the English language.
Descriptions of each linguistic property class are included alongside the dataset release.

Linguistic annotations for an entity include those that are important to handle in a template-based language generation context. For instance in English, location entities have locative preposition annotations while people entities have gender annotations.\footnote{Note that the gender annotations may be sometimes incomplete or inaccurate due to changed state of the world, an annotator mistake, or a lack of standard linguistic handling for gender non-binary persons in certain languages.}
In other languages like French, \emph{all} entities are annotated for grammatical gender, and entities with an article are marked depending on whether its article stays unchanged or gets merged with a preposition (like it would for common nouns).

We use linguists who are native speakers in their corresponding language
to source linguistic
annotations for popular entities. Except for the following eight low-resource languages---Bengali~(bn), Gujarati~(gu),
Kannada~(kn), Malayalam~(ml), Marathi~(mr), Tamil~(ta), Telugu~(te), and Urdu~(ur)---all annotators possess at least a bachelors degree in some branch of linguistics. Linguist annotators' median age ranges from 25 to 35, and they are roughly equally
split between male and female.
Instead of expert linguists---or in addition to them---one may use data mining techniques~\cite{gutman2018crafting}.

\begin{table*}[tb]
\begin{tabular}{lp{11em}p{18em}p{6em}}
\toprule
\textbf{lang} & \textbf{name} & \textbf{signature} & \textbf{semantic type }\\
\midrule
fr & Suisse & \texttt{number:SINGULAR,gender:FEMININE, starts\_w\_vowel:0} & Country \\
\\
de & Champions League & \texttt{number:SINGULAR,gender:FEMININE, article:DEFINITE, locative\_prep:PREP\_IN, directional\_prep:PREP\_NACH} & Event \\
\\
ru & \foreignlanguage{russian}{Саратовские авиалини} &
\texttt{number:PLURAL,casus:NOMINATIVE, allative:PREP\_K, comitative:PREP\_S,topical:PREP\_O,
locative\_prep\_geo:PREP\_V} & Corporation \\
\bottomrule
\end{tabular}
\caption{Examples of CLSE linguistic signatures (truncated for conciseness).%
}
\label{tab:signature}
\end{table*}

We introduce the concept of a \emph{linguistic signature}, which is a linearized string representation of an entity’s linguistic attributes for a specific language. Table~\ref{tab:signature} illustrates some examples of linguistic signatures.

The maximum hypothetical number of distinct linguistic signatures for a language is the Cartesian product of all linguistic features and values for that language. However, not all linguistic signatures are naturally occurring or relevant. For example, mass nouns\footnote{\myurl{thoughtco.com/uncountable-noun-spanish-3079280}} that start with a vowel do not occur in Spanish. Consequently we source \emph{different} entities for different languages
(and different number of them) to ensure linguistic diversity appropriate for each language.

To obtain entities based on linguistic variation, we annotate a large number of entities for each semantic type to create a table of ($language$, $semantic\_type$, $entity\_id$, $name$, $linguistic\_signature$). We group rows in the table by ($language$, $semantic\_type$, $linguistic\_signature$) triples.
The complete corpus covers 34 languages, 74 semantic types, and 222 distinct linguistic signatures.

The full corpus is available at GitHub: \myurl{github.com/google-research-datasets/clse}.

\section{CLSE Case Study: Task-Oriented Dialogs}
\label{sec:sgd}

To demonstrate how one may use this corpus, we consider the problem of NLG for task-oriented dialog
systems~\cite{wen2015semantically}.
Unlike open-domain chit-chat systems, the natural language interface of virtual assistants such as Amazon Alexa,
Apple Siri, or Google Assistant is highly task-oriented.
Users often interact with their virtual assistants to accomplish a specific action,
like finding flights, booking restaurants, buying tickets, etc.

In a task-oriented dialogue, the conversation between the user and the assistant is tracked by a dialog manager that uses a dialog state,
a summary of the entire conversation up to the current turn (see, e.g., \cite{pieraccini2021ai}). The dialog state consists of slots and values related to the specific intents, services, and actions in question. The assistant uses the dialog state to 1) invoke external APIs with appropriate parameter values, as specified by the user over the dialog history, and 2) generate next actions to continue the dialog, for example soliciting for more information from the user or confirming the user's intent or request \cite{aliannejadi2021building}. Finally, 3) the selected dialog actions along with structured data is used to generate a new utterance to respond back to the user (the NLG task).

\begin{figure*}[tb]
\centering
\includegraphics[width=0.9\linewidth]{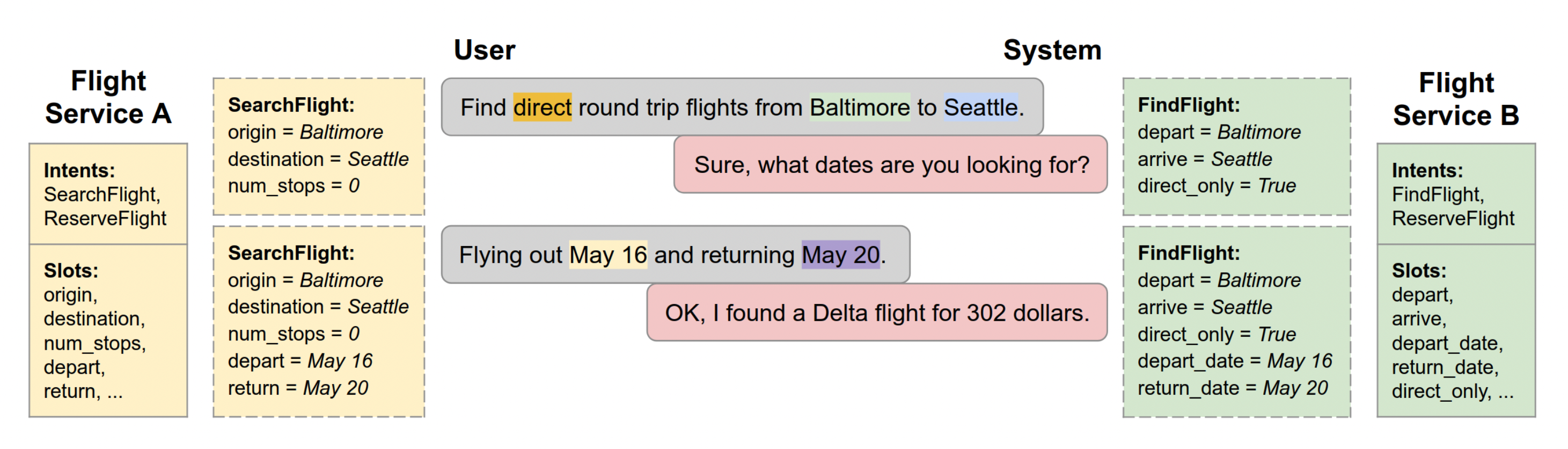}
\caption{An example of a conversation from a schema-guided dialog~\cite{rastogi2020towards}.  The predicted dialogue state (shown with dashed edges) for the first two user turns for an example dialogue, showing the active intent and slot assignments, with two related annotation schemas. Note that the dialogue state representation is conditioned on the schema under consideration, which is provided as input, as are the user and system utterances.}
\label{fig:sgd_example}
\end{figure*}

\begin{table}[tb]
\begin{tabular}{lcc}
\toprule
\textbf{} & \textbf{Wikidata} & \textbf{CLSE} \\
\midrule
Total \# of arguments       & 20 & 17 \\
Output is missing "the"     & 1  & 6  \\
Wrong preposition "in"      & 0  & 3  \\
Grammatical outputs         & 95\%& 47\%  \\
\bottomrule
\end{tabular}
\caption{Comparing CLSE and WikiData as sources of arguments for testing an NLG system.}
\label{tab:wikidata}
\end{table}

Task-oriented NLG datasets like MultiWOZ~\cite{budzianowski2018multiwoz} and SGD~\cite{rastogi2020towards} are designed to be balanced with respect to the number of turns, intent and slot usage, etc., without much focus on the \emph{linguistic} properties of incoming parameter values.
For the sake of illustration, let us consider the following simplistic NLG system.
It receives a \texttt{location} argument and returns the following templated response:

\code{EMNLP will be held in \$\{location\}.}

\noindent Without consulting a linguist or a native speaker, one can try exposing the issues of this template by substituting
different argument values for \texttt{location}.
The CLSE has 17 entries of type \texttt{AdministrativeArea} with unique linguistic
signatures in English. Note that some signatures may only differ in ways that are inconsequential
to the above templates, but the hope is still that they will be more diverse than a general-purpose list.
For comparison, we consider a general-purpose list of 20 entities of type "administrative territorial entity" (Q56061) 
from Wikidata~\cite{vrandevcic2014wikidata}.\footnote{SPARQL query: \myurl{pastebin.com/KBk17G5k}.}
Table~\ref{tab:wikidata} summarizes the errors using entity arguments from Wikidata and the CLSE.
Wikidata exposes only one potential issue: a missing determiner in front of "Bahamas."
Even worse, as of 2022-09-09, our SPARQL query returns the surface form as "The Bahamas," which means that
if we were to evaluate our NLG system purely based on the realized output texts, we may declare our system to have 100\% fluency, while using the entities from CLSE we immediately see that the system is linguistically brittle, and there
is more than 50\% fluency headroom (one sentence has two mistakes).

Factual accuracy mistakes are also unforgivable when it comes to virtual assistants. \citet{kale2020machine} point out delexicalized models as a less error-prone alternative to lexicalized models. In the delexicalized setting, models are trained to produce output text with placeholders, which are filled in via a separate lexicalization step (usually naive string substitution). The semantic accuracy of delexicalized models tends to be far ahead of their lexicalized counterparts, especially in the presence of slot values not seen during training. However, delexicalization and other copy-based methods are more grammatically deficient in the presence of linguistic phenomena such as morphological inflection (changing surface form of a word depending on its function in a sentence; see, e.g., Table~\ref{tab:morph_inflection}).
This makes a naive delexialization approach suboptimal for highly inflected languages~\cite{cs_restaurants},
a claim that we will also test below.

\section{SGD-CLSE: Generating Linguistically Diverse Data Using the CLSE}\label{sec:sgd_clse}

We perform our experiments on the Schema-Guided Dialog dataset from DCT8~\cite{rastogi2020towards}.
To emulate a data-to-text setup, we pair each utterance (text) with that utterance’s acts, services, and slots. An example of such a pair can be seen in Table~\ref{tab:d2t_example}.
Since the scope of our experiments is to focus on linguistic robustness with regard to entities,
we only look at SYSTEM utterances, and ignore all examples that don’t use any entity slots.
This results in 707 $(\mathit{service\_name}, \mathit{action}, \mathit{slot\_names})$ triples
from the SGD train set and 23 triples from the SGD test set.

\begin{table*}[tb]
\begin{tabular}{p{25em}p{14em}}
\toprule
\textbf{input (structured data)} & \textbf{output (utterance)} \\
\midrule
\textbf{service\_name:} "Restaurants\_1" & \multirow{3}{14em}{"How would you like Bazille, which is situated in San Jose?"} \\
\textbf{actions:}  ["OFFER\_restaurant", "OFFER\_city"] & \\
\textbf{slots:} \{restaurant: "Bazille", city: “San Jose”\} & \\
\bottomrule
\end{tabular}
\caption{An example of structured data input and free text output for an SGD dialog utterance.}
\label{tab:d2t_example}
\end{table*}

\begin{figure*}[tb]
\centering
\includegraphics[width=1.0\linewidth]{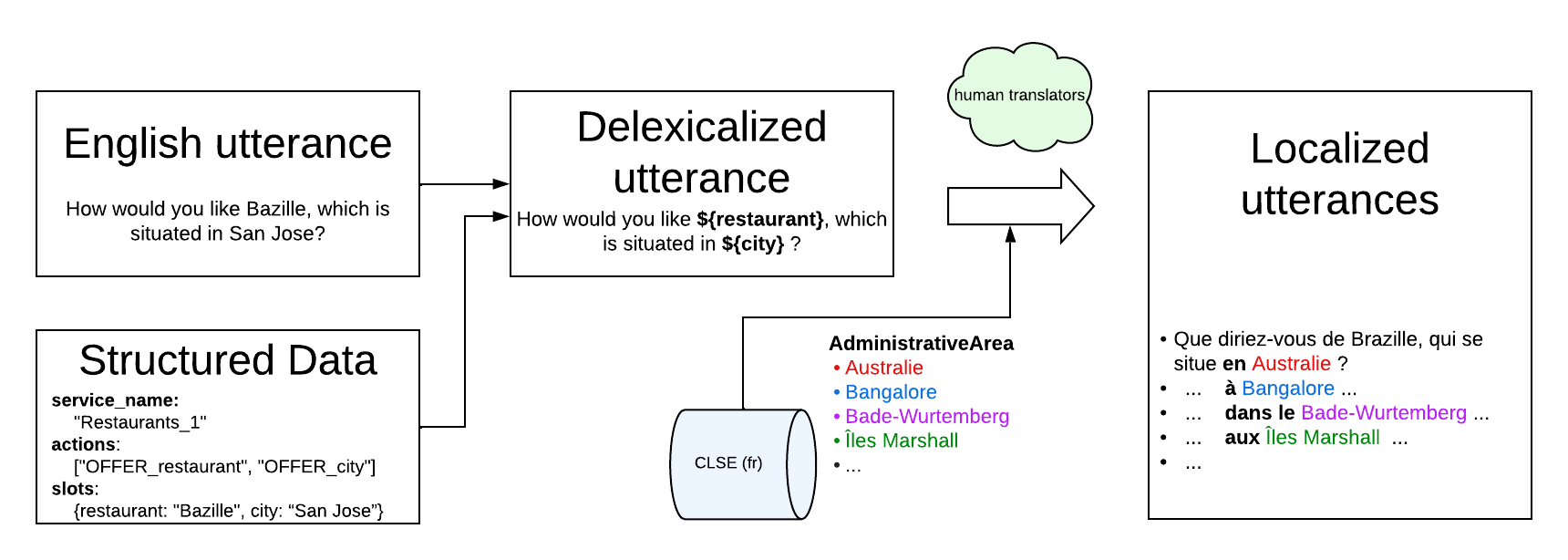}
\caption{Algorithm for creating the SGD-CLSE dataset (example for French).}
\label{fig:sgd_clse}
\end{figure*}

Owing to the lack of noun inflections in English, we can also produce a delexicalized form of the utterance
by searching for and stubbing out the argument values within.
E.g., for the utterance in Table~\ref{tab:d2t_example} we would get
\texttt{"How would you like \$\{restaurant\}, which is situated in \$\{city\}?"}. The delexicalized form gives us an English template that allows us to substitute other values for the placeholders to produce new realistic utterances, enabling data augmentation methods like the one described by \citet{kale2021mixout}. 

In our case study, we assume that we already have a system that can produce English utterances---arguably a much easier task given the simplicity of English grammar---and instead focus on localization: generating fluent and grammatical output in 
a target language.
In this setup we get the following as input: dialog state, structured data, English text, and target language. We focus on three languages: French, Russian, and Marathi.
The motivation for choosing these languages is to have a widely studied high-resource language (French),
a highly inflected language (Russian), and a low-resource language (Marathi).

The process of creating the SGD-CLSE dataset, leveraging the CLSE, is as follows.
1) for every $(\mathit{service\_name}, \mathit{action}, \mathit{slot\_names})$ triple,
we generate argument values by randomly sampling slot values from the CLSE for
a specific \emph{target} language: French, Russian, or Marathi respectively.
This requires us to establish a mapping from each entity-related slot name to a semantic type. For example, $city$ slots in the SGD schemas would be mapped to the \href{https://schema.org/City}{City} schema.org type in the CLSE corpus. %
2) we substitute these new argument values into the English delexicalized utterances (templates) from SGD.
The goal of entity re-sampling using the CLSE is to generate as linguistically diverse data as possible, for maximum linguistic coverage in the \emph{target} language.\footnote{It is possible that substituting new arguments could break the fluency of the example in the source language (English, in our case).
However, we found that imperfect substitutions in the source language do not influence the quality of human translations,
produced by the process we describe further.}
3) Finally, we use professional translators to translate those utterances into the corresponding languages. Figure~\ref{fig:sgd_clse} illustrates the algorithm.

We use the aforementioned process with three examples per $(\mathit{service}, \mathit{action}, \mathit{slot\_names})$ triple to generate language-specific linguistically diverse examples in French, Russian, and Marathi\footnote{It is worth noting that the quality bar for the CLSE annotations was higher in Hindi compared to the other Indic languages. While it might still be more appropriate to use the Marathi section of the CLSE for other applications, we empirically found that the Hindi section yields SGD-based realizations of higher quality, while maintaining linguistic diversity. This is not entirely surprising, since the languages are close geographically (both are spoken in India) and linguistically (both are Indo-Aryan languages stemming from Sanskrit). We therefore use the CLSE's Hindi entities as a proxy for Marathi ones.} on the train partition of the SGD dataset. We then use two of these examples for \textbf{train} and one
for what we call "within-triple test" split (\textbf{wt-test}).\footnote{In cases where we were only able to obtain two distinct examples,
we use one of them for train and one for wt-test. In case of a single example, the triple is discarded completely.}
Unlike the SGD dev and test splits, the wt-test split has a particular focus on linguistics
as opposed to generalization across dialog states:
it contains states that were seen during training, but with a linguistically diverse set of slot values.
In addition to train and wt-test sets, we also use a sample from the original
SGD \textbf{dev} and \textbf{test} sets translated into the target languages.
Table~\ref{tab:dataset} has dataset statistics.

\begin{table}[tb]
\begin{tabular}{lcccc}
\toprule
\textbf{lang} & \textbf{train} & \textbf{wt-test} & \textbf{SGD test} & \textbf{SGD dev} \\
\midrule
fr & 451 & 233 & 277 & 187 \\
ru & 451 & 236 & 277 & 187 \\
mr & 451 & 234 & 277 & 187 \\
\bottomrule
\end{tabular}
\caption{Number of items in different partitions of the SGD-CLSE dataset.}
\label{tab:dataset}
\end{table}

\section{Experimental Setup}
\label{sec:experiment}
Below we describe our setup to evaluate different NLG models using the SGD-CLSE dataset.

\subsection{Models for Comparison}
\label{sec:models}

\begin{itemize}
    \item \textbf{nmt}: An out-of-the-box machine translation model with English plain text as input. We used the \texttt{GOOGLETRANSLATE} function of Google Sheets.\footnote{\myurl{support.google.com/docs/answer/3093331}; translations for the wt-test were obtained on 2022-01-28, for SGD test/dev: on 2022-02-02.}
    \item \textbf{d2t}: A data-to-text model fine-tuned on the available train set, with best checkpoint picked on the dev set. We use a pretrained \textit{mT5 xxl} model~\cite{xue2020mt5} as a basis for our fine-tuning.\footnote{Our early experiments with \textit{mT5 base} gave substantially lower BLEU scores, suggesting that the bigger models yield stronger baselines. We use \textit{xxl} models going forward.}
    \item \textbf{tmpl}: Collect translated templates (delexicalized utterances) for the train set. Slot values are plugged in verbatim without any morphological inflection. Note that for most triples we have two different translations
    available for train (Table~\ref{tab:dataset}). We only use one of them (picked at random) but also report
    confidence intervals based on possibly picking different translations as bases for
    delexicalization.\footnote{We use a simple bootstrap procedure: pick one of the two translations
    for each triple by flipping a fair coin. The process is repeated 1000 times and the 5th and 95th percentiles
    are reported as the confidence interval bounds.}
    \item \textbf{tmpl+G}: Same as above with a grammatical error correction~(GEC) model applied on top of the template output. We use \textit{gT5 xxl} model by~\citet{rothe2021simple}.
\end{itemize}

\begin{figure*}[tb]
\centering
\includegraphics[width=1.0\linewidth]{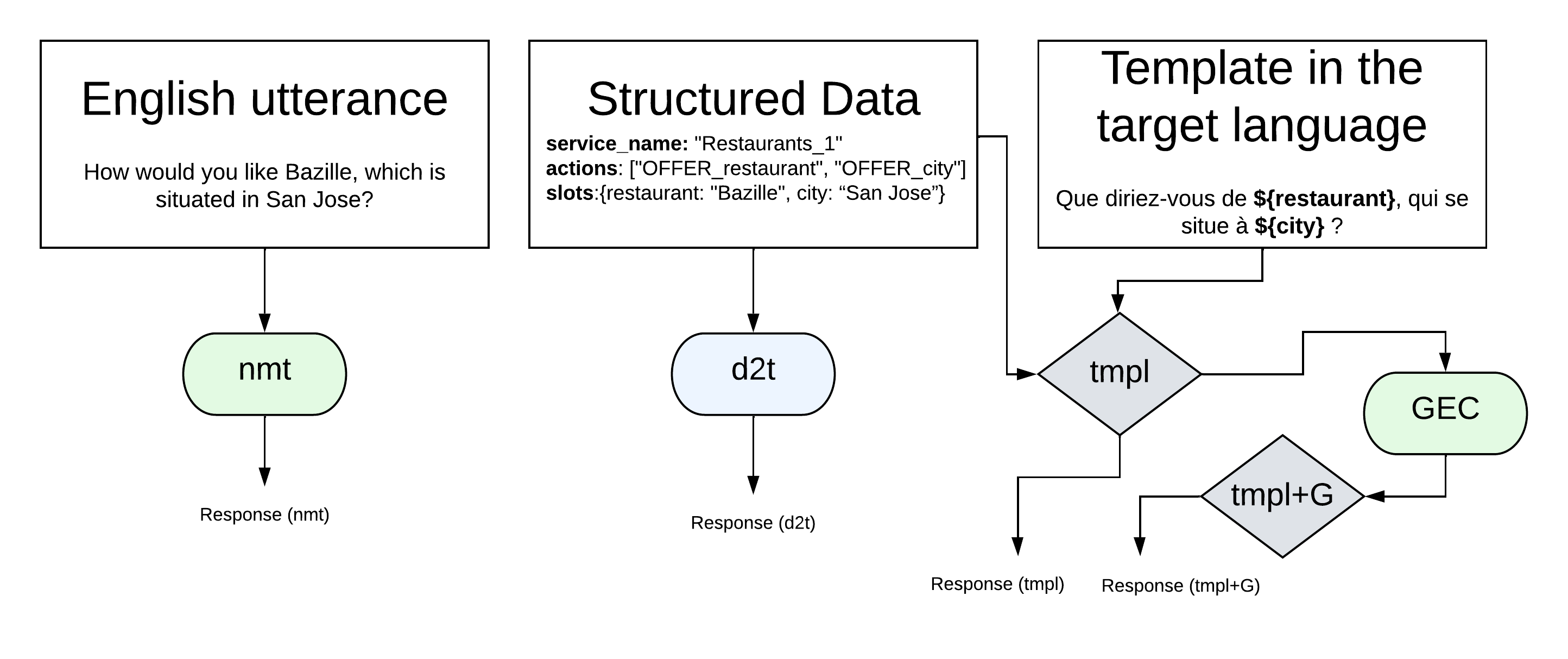}
\caption{Input data flow for various baseline NLG systems. \textbf{GEC} and \textbf{nmt}
(green)
are off-the-shelf models, while \textbf{d2t} (azure) is a model fine-tuned on given data.
\textbf{tmpl} and \textbf{tmpl+G} are simple pipeline algorithms (rhombus shape).}
\label{fig:clse_sgd_baselines}
\end{figure*}

Figure~\ref{fig:clse_sgd_baselines} illustrates which parts of the input data are consumed by which models.

\subsection{Training Details}
For fine-tuning the \textbf{d2t} model we use a batch size of 64 and fine-tuned on TPU for 5'000 steps with a learning rate of $10^{-3}$. We trained the models independently for each language and picked the stopping point based on the corresponding dev set. The model has 13B parameters. 

\subsection{Metrics}
We use BLEU\footnote{\faGithub\ \myurl{github.com/tuetschek/e2e-metrics}.}~\cite{papineni2002bleu}
and BLEURT\footnote{BLEURT-20 \faGithub\  \myurl{github.com/google-research/bleurt}.}~\cite{sellam2020bleurt} as our automatic metrics.
For human evaluation, we assess fluency and factual accuracy.
Table~\ref{tab:agreement} summarizes agreement between raters.

\paragraph{Accuracy:} Human raters are shown an NLG system's output in the target language as well as the English text as a reference. They are instructed to mark the NLG system output as inaccurate if any information contradicts the English reference. This effectively catches errors due to hallucinations, incorrect grounding etc. Each example is rated by three raters. We take the average of the accuracy scores ($1$ for accurate, $0$ otherwise).

\paragraph{Fluency:} We ask the raters how grammatical an NLG system's output sounds on a 1 to 5 Likert scale, with 5 being the highest score. Again, each example is rated by three raters. We average the scores across all the ratings to get the fluency score.

\begin{table}[tb]
\begin{tabular}{llll}
\toprule
& \textbf{French} & \textbf{Russian} & \textbf{Marathi} \\
\midrule
Accuracy & 0.77 (s) & 0.69 (s) & 0.63 (s) \\ 
Fluency  & 0.43 (m) & 0.51 (m) & 0.35 (f) \\
\bottomrule
\end{tabular}
\caption{Kappa coefficient~\cite{cohen1960coefficient} inter-rater agreement.
Values 0.21--0.40 are considered fair (f), 0.41--0.60 moderate (m) and 0.61--0.80 substantial (s).}
\label{tab:agreement}
\end{table}

\section{Results}
\label{sec:results}

We split this section into two. First, we look at the results on the unseen test set. These contain dialog
situations, $(\mathit{service\_name}, \mathit{action}, \mathit{slot\_names})$,  which the systems did not see during training. We then move to the wt-test split, where we expect the gap to the human-written responses to be smaller.

\subsection{Unseen Test Set}

Table~\ref{tab:sgd_test_results} contains the summary of the results. 
Naturally, the template-based approaches are not able to generalize to these situations, so they are not
included in the results here.

\begin{table}[tb]
\begin{tabular}{ll@{~~}c@{~~}c@{~~}ll}
\toprule
 & & \textbf{BLEU} & \textbf{BLEURT} & \textbf{acc.} & \textbf{fl.} \\
\midrule
\multirow{3}*{fr} & d2t   & 0.14 & 0.39 & 0.78 & 4.58 \\
                  & nmt   & 0.32 & 0.62 & 0.96$^\blacktriangle$ & 4.43$^\blacktriangledown$ \\
                  & human & -    & -    & 0.98  & 4.78$^\blacktriangle$ \\
\midrule
\multirow{3}*{ru} & d2t & 0.15 & 0.50 & 0.50 & 4.35 \\
                  & nmt & 0.16 & 0.57 & 0.79$^\blacktriangle$ & 3.70$^\blacktriangledown$ \\
                   & human &  -  & -  & 0.96$^\blacktriangle$ & 4.89$^\blacktriangle$ \\
\midrule
\multirow{3}*{mr} & d2t & 0.07 & 0.60 & 0.41 & 3.50 \\
                  & nmt & 0.12 & 0.71 & 0.77$^\blacktriangle$ & 4.15$^\blacktriangle$ \\
                  & human & -  & -    & 0.92$^\blacktriangle$ & 4.71$^\blacktriangle$ \\
\bottomrule
\end{tabular}
\caption{Results of the NLG models on the SGD test set (unseen triples).
For human scores---accuracy (acc.) and fluency (fl.)---we also mark whether those are statistically
significantly different from the \emph{previous row} using paired t-test:
$^\blacktriangle$ or $^\blacktriangledown$ denote significant difference at $p = 0.01$,
$^\vartriangle$ or $^\triangledown$ --- at $p = 0.05$ respectively.}
\label{tab:sgd_test_results}
\end{table}

We could verify that all NLG systems are noticeably (and significantly) behind human translations in terms of fluency.
In terms of accuracy, on the other hand, the \textbf{nmt} baseline is \emph{in}significantly
below the human bar for French, suggesting that we do benefit from the high-resource nature of the language.
Another observation is that, while the \textbf{nmt} baseline appears to outperform \textbf{d2t} on all dimensions
for a low-resource Marathi, the picture is less clear for French and Russian.
There we see that the \textbf{d2t} actually scores significantly \emph{lower} than \textbf{nmt} in terms of accuracy
but \emph{higher} in terms of fluency.
We believe that the accuracy gap could be lowered with more training data, which
we leave for future work to investigate.

\subsection{Within-Triples Test}
\label{sec:wt_test_results}

\begin{table}[tb]
\begin{tabular}{l@{~~}l@{~~}c@{~}c@{~~}p{3em}@{~~}p{3em}}
\toprule
&  & \textbf{BLEU} & \textbf{BLEURT} & \textbf{acc.} & \textbf{fl.} \\
\midrule
\multirow{5}*{fr} & nmt & 0.39 & 0.65 & \,0.90 & \,4.19 \\
                 & tmpl & 0.46 & 0.70 & \,0.88 \tiny{[0.87, 0.90]} & \,4.31$^\vartriangle$ \tiny{[4.26, 4.34]} \\
               & tmpl+G & 0.47 & 0.71 & \,0.91$^\blacktriangle$ \tiny{[0.88, 0.91]} & \,4.48$^\blacktriangle$ \tiny{[4.43, 4.49]} \\
                  & d2t & 0.41 & 0.65 & \,0.91 & \,4.64$^\blacktriangle$ \\
                & human & -    & -    & \,0.94 & \,4.63 \\
\midrule
\multirow{5}*{ru} & nmt & 0.15 & 0.57 & \,0.71 & \,3.37 \\
                  & d2t & 0.37 & 0.66 & \,0.72$^*$ & \,4.51$^{\blacktriangle*}$ \\
                 & tmpl & 0.41 & 0.72 & \,0.79$^\blacktriangle$ \tiny{[0.79, 0.83]} & \,3.95$^\blacktriangledown$ \tiny{[3.93, 4.03]} \\
               & tmpl+G & 0.44 & 0.74 & \,0.79 \tiny{[0.77, 0.81]} & \,4.21$^\blacktriangle$ \tiny{[4.17, 4.28]} \\
                & human & -    & -    & \,0.87$^\blacktriangle$ & \,4.62$^\blacktriangle$ \\
\midrule
\multirow{5}*{mr} & nmt & 0.15 & 0.69 & \,0.72 & \,3.73 \\
                  & d2t & 0.33 & 0.69 & \,0.66 & \,4.05$^\blacktriangle$ \\
                 & tmpl & 0.51 & 0.78 & \,0.83$^\blacktriangle$ \tiny{[0.80, 0.83]} & \,4.32$^\blacktriangle$ \tiny{[4.27, 4.34]} \\
               & tmpl+G & 0.51 & 0.78 & \,0.82 \tiny{[0.79, 0.81]} & \,4.29 \tiny{[4.26, 4.33]} \\
                & human & -    & -    & \,0.92$^\blacktriangle$ & \,4.50$^\blacktriangle$ \\
\bottomrule
\end{tabular}
\caption{Results of the baseline NLG systems on the wt-test set (seen triples).
For human scores---accuracy (acc.) and fluency (fl.)---we also mark whether those are statistically
significantly different from the \emph{previous row} using paired t-test:
$^\blacktriangle$ or $^\blacktriangledown$ denote significant difference at $p = 0.01$,
$^\vartriangle$ or $^\triangledown$ --- at $p = 0.05$ respectively.
The square brackets for template-based approaches denote  95\% confidence intervals obtained
using the bootstrap procedure described in Section~\ref{sec:models}.
\\[1mm]
\footnotesize{($^*$) Human scores for d2t in Russian come from a different rater population and may not be directly comparable.}
}
\label{tab:wt_test_results}
\end{table}

Table~\ref{tab:wt_test_results} contains a summary of performance of different
NLG systems while Table~\ref{tab:examples} in Appendix~\ref{app:examples} contains example outputs.
We see notable quality gains of the \textbf{d2t} or template-based approaches compared to the
off-the-shelf \textbf{nmt} system. %

The grammatical error correction model (\textbf{tmpl+G} baseline), appears to improve the results on top of the pure template-based \textbf{tmpl} baseline
for high-resource languages. The gain is higher for Russian, a highly inflected language. No measurable effect is reported on Marathi, a low resource language,
suggesting that the grammar error correction model
itself may not be of sufficient quality. The fluency \textbf{tmpl+G} achieves for Russian is still significantly lower than that of \textbf{d2t},
but comes with a significantly higher accuracy (inasmuch as we can compare them given that the fluency and accuracy scores for \textbf{d2t}
come from a different rater pool).

It is interesting to note that the \textbf{d2t} model appears to get higher human scores for French than the other NLG systems,
but scores lower on automatic metrics. In fact, the human scores
are not statistically different from that of the human baseline. Upon closer examination, however, we see that this model still frequently makes grave mistakes.
One explanation for this could be that the humans have a tolerance for hallucinations or missing facts when there are many of them presented in the same utterance.
This is consistent with previous findings of~\citet{freitag2021experts}.
Appendix~\ref{app:examples} shows examples where the automatic metrics are right to penalize the \textbf{d2t} model
for accuracy while raters completely miss it.
While prior work such as \cite{pagnoni-etal-2021-understanding,honovich2022true} studies factual consistency in English, further work on evaluating factual consistency of \emph{localization} approaches is needed.

\medskip\noindent%
To summarize the above results, we can conclude that
there is still a gap between outputs generated by our baseline NLG systems and responses written by humans,
especially for lower-resource languages like Marathi or morphologically complex languages such as Russian. Among baseline systems,
the template-based \textbf{tmpl}/\textbf{tmpl+G} approaches, when available, clearly outperform the \textbf{d2t} model for Marathi. The results of the automatic metrics in other languages suggest the same conclusion, though human scores are not always consistent with that.
We attribute this inconsistency to the fact that the \textbf{d2t} system
can generate seemingly good responses that, while missing or introducing facts, manage to convince human raters of their accuracy
(see Appendix~\ref{app:examples}).

Even with this caveat, we can see that accuracy scores of the template-based approach (\textbf{tmpl+G})
is the same or higher than that of \textbf{nmt} or \textbf{d2t} baselines,
suggesting that the factual accuracy is a fundamental weakness of neural approaches.
At the same time, the template-based approaches still suffer from fluency issues, even with the grammatical error correction model applied.
We hypothesize that the main reason is that the task of correcting mistakes in the template-based approach
does not exactly map to the grammatical error correction task.
There are types of mistakes we see here that humans rarely make: e.g., inserting determiners
in front of popular city names like "London" or "Paris" or confusing dative for nominative (dative and accusative are more commonly confused by humans).
This suggests that developing a dedicated grammar model for NLG may be helpful.

\section{Conclusion and Discussion}
Building a natural language generation system that can handle a broad diversity of entities with varying linguistic phenomena remains an open challenge. With the CLSE, any schema-informed NLG datasets can use techniques
described in Sections~\ref{sec:sgd}~and~\ref{sec:sgd_clse} to produce better linguistically represented data, and measure metrics on that data to draw more linguistically thoughtful conclusions.
The space of possible inputs that NLG systems will be expected to handle may be highly unconstrained,
and designing solutions that are linguistically robust, and defensibly so, is an ambitious and worthwhile pursuit.

Our results in Section~\ref{sec:results} establish an evaluation procedure probing NLG systems
for their linguistic capabilities. We also evaluate four baselines using this procedure
and conclude that none of them is at the human level yet.
Improving upon these baselines can be approached from two sides: either 1) improving factual accuracy
and reducing hallucinations of a purely neural data-to-text approach or 2) improving the quality of
grammatical error correction applied to a template-based approach.

Beyond NLG for virtual assistants, the CLSE data may also be used for other NLG tasks.
E.g., we can use it to augment machine translation datasets by swapping named entities 
to further probe the fluency of generated translations.

\medskip\noindent%
The full CLSE dataset is openly available at \myurl{github.com/google-research-datasets/clse}.

\section{Ethical Considerations}
Releasing a dataset in multiple languages, including
several low-resource ones, would allow to push the state of research
in non-English NLG\@. While it is not impossible that this dataset might be used
for building ML models for malicious applications, we believe
it will be widely used for public good and will be a net positive contribution to society.

Crowd-sourced annotations were collected using a proprietary crowd-sourcing
platform.
Workers were paid at least 50\% more than the minimum hourly wage.
No information about the workers will be released and
worker IDs are anonymized.

\section{Limitations}
The dataset we release and the experiments we conduct have number of limitations.
Firstly, there are other linguistically significant phenomena that arise from non-entities like verbs and numbers that the CLSE does not include.
Then we only cover 34 languages---small in comparison to, say, Wikipedia or Common Crawl datasets~\cite{conneau2019cross,xue2020mt5}
covering 100+ languages. Moreover, the quality of annotations for low-resource languages is lower due to limited linguist resources.
Not only the quality, but the quantity of annotated entities varies greatly across languages (Table~\ref{tab:clse_stats}),
either due to some languages having fewer linguistic signatures, or annotator resource constraints.
The experiments we conduct with SGD cover only a subset of CLSE in terms of languages and semantic types.
The SGD-based dataset we used for our case study is rather small (limited by the human translation budget),
and human ratings were not free of biases (e.g., humans were often more forgiving for accuracy mistakes than the automatic metrics were).
Our experiments do not include an NMT model fine-tuned on the human translations---a common domain adaptation technique \cite{luong2015stanford,neubig2018rapid,bapna2019simple}---in favor of a direct data-to-text model.
Finally, we used the \textit{xxl} versions of the models, which require significant computational resources to train and run.

\section{Acknowledgements}
The authors would like to thank Ariel Gutman, Eric Malmi, Jelena \v{C}uklina, and anonymous reviewers
for their comments and suggestions on how to improve the paper.
We would also like to thank Katie Vadella for her help annotating the linguistic phenomena as well
as the Google Assistant Linguist Team, Google AI Ethics Team, Google Search Data Team
for shaping and supporting the data release.

All content represents the opinion of the authors, which is not necessarily shared or endorsed by their respective employers.

\bibliography{custom}
\bibliographystyle{acl_natbib}

\clearpage
\appendix

\onecolumn
\section{CLSE Dataset Statistics}\label{sec:dataset_stats}

Table~\ref{tab:clse_stats} contains statistics of the CLSE dataset per language. We employ commonly used
2-letter language tags (ISO-639-1), except for "cmn-CN" for Chinese Mandarin written using the simplified script, "cmn-TW" for Chinese Mandarin written with traditional script, and "yue" for the Cantonese Chinese (written using traditional script).
We do not use the Chinese macro-tag "zh" because it is important to distinguish the above locales for NLG purposes.

\begin{table*}[h]
\begin{tabular}{lr@{~~}r@{~~}r@{~~}r@{~~}r@{~~}r@{~~}r@{~~}r@{~~}r@{~~}r@{~~}rr}
\toprule
language &    ar &    bn &  cmn-CN &  cmn-TW &    cs &    da &    de &     en &    es &     fr &    gu \\
\# unique entities  &  899 &  721 &     530 &     529 &  1238 &  1286 &  2922 &  4076 &  3181 &  4312 &  798 \\
\# ling. attributes &   21 &    6 &       5 &       6 &    20 &    23 &    37 &    32 &    26 &    26 &    6 \\
\midrule
language &    hi &    id &    it &    ja &   jv &    kn &    ko &    ml &    mr &    nl &    no \\
\# unique entities  &  950 &  705 &  2510 &  1063 &  55 &  849 &  885 &  888 &  924 &  1049 &  1237 \\
\# ling. attributes &   18 &    9 &    41 &    47 &   2 &    6 &    9 &    6 &    6 &    20 &    23 \\
\midrule
language &    pl &    pt &     ru &   su &    sv &    ta &    te &    th &    tr &    ur &    vi &   yue \\
\# unique entities  &  1606 &  2464 &  3039 &  29 &  1309 &  891 &  885 &  724 &  1262 &  883 &  612 &  551 \\
\# ling. attributes &    30 &    25 &    31 &   2 &    19 &    6 &    6 &    9 &    10 &    6 &    8 &   19 \\
\bottomrule
\end{tabular}
\caption{Per language statistics of the CLSE dataset. The number of annotated entities across different languages varies greatly, either due to fewer linguistic signatures applicable to a given language
or annotator resource constraints.}
\label{tab:clse_stats}
\end{table*}

The complete corpus covers 34 languages, 74 semantic types, and 222 distinct linguistic signatures.
Table~\ref{tab:semantic_types} contains a full list of the semantic types present in the dataset.

\LTcapwidth=0.98\textwidth
\begin{small}
\begin{longtable}{p{.27\textwidth}p{.6\textwidth}}
\toprule
 
\textbf{Semantic Type} & \textbf{Description} \\
\endfirsthead

\multicolumn{2}{l}
{{\bfseries \tablename\ \thetable{} -- continued from previous page}} \\
\toprule
\textbf{Semantic Type} & \textbf{Description} \\ \midrule 
\endhead

\endfoot

\endlastfoot

\midrule
AdministrativeArea &
  A geographical region, typically under the jurisdiction of a particular government.  \\
\midrule
Airline &
  An organization that provides flights for passengers.  \\
\midrule
Airport &
  An airport.  \\
\midrule
AmusementPark &
  An amusement park.  \\
\midrule
Article &
  An article, such as a news article or piece of investigative report. Newspapers and magazines have articles of many different types and this is intended to cover them all. \\
\midrule
BodyOfWater &
  A body of water, such as a sea, ocean, or lake.  \\
\midrule
Book &
  A book.  \\
\midrule
BookSeries &
  A series of books. Included books can be indicated with the hasPart property.  \\
\midrule
Brand &
  A brand is a name used by an organization or business person for labeling a product, product group, or similar.  \\
\midrule
Bridge &
  A bridge.  \\
\midrule
BroadcastChannel &
  A unique instance of a BroadcastService on a CableOrSatelliteService lineup.  \\
\midrule
BroadcastService &
  A delivery service through which content is provided via broadcast over the air or online.  \\
\midrule
BusStation &
  A bus station. \\
\midrule
CableOrSatelliteService &
  A service which provides access to media programming like TV or radio. Access may be via cable or satellite.  \\
\midrule
Cemetery &
  A graveyard.  \\
\midrule
City &
  A city or town.  \\
\midrule
CivicStructure &
  A public structure, such as a town hall or concert hall.  \\
\midrule
CollegeOrUniversity &
  A college, university, or other third-level educational institution.  \\
\midrule
Continent &
  One of the continents (for example, Europe or Africa).  \\
\midrule
Corporation &
  Organization: A business corporation.  \\
\midrule
Country &
  A country.  \\
\midrule
CreativeWork &
  The most generic kind of creative work, including books, movies, photographs, software programs, etc.  \\
\midrule
DefenceEstablishment &
  A defence establishment, such as an army or navy base.  \\
\midrule
Diet &
  A strategy of regulating the intake of food to achieve or maintain a specific health-related goal.  \\
\midrule
EducationalOrganization &
  An educational organization.  \\
\midrule
Event &
  An event happening at a certain time and location, such as a concert, lecture, or festival. Ticketing information may be added via the offers property. Repeated events may be structured as separate Event objects.  \\
\midrule
Game &
  The Game type represents things which are games. These are typically rule-governed recreational activities, e.g. role-playing games in which players assume the role of characters in a fictional setting.  \\
\midrule
GovernmentOrganization &
  A service provided by a government organization, e.g. food stamps, veterans benefits, etc.  \\
\midrule
GovernmentService &
  A service provided by a government organization, e.g. food stamps, veterans benefits, etc.  \\
\midrule
Hospital &
  A hospital.  \\
\midrule
ItemList &
  A list of items of any sort—for example, Top 10 Movies About Weathermen, or Top 100 Party Songs. Not to be confused with HTML lists, which are often used only for formatting.  \\
\midrule
LakeBodyOfWater &
  A lake (for example, Lake Pontrachain). \\
\midrule
LandmarksOrHistoricalBuildings &
  An historical landmark or building.  \\
\midrule
LocalBusiness &
  A particular physical business or branch of an organization. Examples of LocalBusiness include a restaurant, a particular branch of a restaurant chain, a branch of a bank, a medical practice, a club, a bowling alley, etc. \\
\midrule
LodgingBusiness &
  A lodging business, such as a motel, hotel, or inn.  \\
\midrule
MobileApplication &
  A software application designed specifically to work well on a mobile device such as a telephone. \\
\midrule
Mountain &
  A mountain, like Mount Whitney or Mount Everest. \\
\midrule
Movie &
  A movie. \\
\midrule
MovieSeries &
  A series of movies. Included movies can be indicated with the hasPart property. \\
\midrule
MovieTheater &
  A movie theater. \\
\midrule
Museum &
  A museum.  \\
\midrule
MusicAlbum &
  A collection of music tracks. \\
\midrule
MusicComposition &
  A musical composition.  \\
\midrule
MusicGroup &
  A musical group, such as a band, an orchestra, or a choir. Can also be a solo musician. \\
\midrule
MusicRecording &
  A music recording (track), usually a single song. \\
\midrule
MusicVenue &
  A music venue.  \\
\midrule
Organization &
  An organization such as a school, NGO, corporation, club, etc. \\
\midrule
Periodical &
  A publication in any medium issued in successive parts bearing numerical or chronological designations and intended, such as a magazine, scholarly journal, or newspaper to continue indefinitely. \\
\midrule
Person &
  A person (alive, dead, undead, or fictional). \\
\midrule
Place &
  Entities that have a somewhat fixed, physical extension. \\
\midrule
PlaceOfWorship &
  Place of worship, such as a church, synagogue, or mosque. \\
\midrule
Product &
  Any offered product or service. For example: a pair of shoes; a concert ticket; the rental of a car; a haircut; or an episode of a TV show streamed online. \\
\midrule
ProductModel &
  A datasheet or vendor specification of a product (in the sense of a prototypical description). \\
\midrule
RadioStation &
  A radio station. \\
\midrule
Restaurant &
  A restaurant. \\
\midrule
RiverBodyOfWater &
  A river (for example, the broad majestic Shannon). \\
\midrule
School &
  A school.  \\
\midrule
SingleFamilyResidence &
  Residence type: Single-family home. \\
\midrule
SoftwareApplication &
  A software application. \\
\midrule
SportsOrganization &
  Represents the collection of all sports organizations, including sports teams, governing bodies, and sports associations. \\
\midrule
SportsTeam &
  Organization: Sports team.  \\
\midrule
StadiumOrArena &
  A stadium.  \\
\midrule
TVSeason &
  Season dedicated to TV broadcast and associated online delivery.  \\
\midrule
TVSeries &
  CreativeWorkSeries dedicated to TV broadcast and associated online delivery.  \\
\midrule
TelevisionChannel &
  A unique instance of a television BroadcastService on a CableOrSatelliteService lineup.  \\
\midrule
TheaterGroup &
  A theater group or company, for example, the Royal Shakespeare Company or Druid Theatre.  \\
\midrule
TouristAttraction &
  A tourist attraction. In principle any Thing can be a TouristAttraction, from a Mountain and LandmarksOrHistoricalBuildings to a LocalBusiness. This Type can be used on its own to describe a general TouristAttraction, or be used as an additionalType to add tourist attraction properties to any other type.  \\
\midrule
VideoGame &
  A video game is an electronic game that involves human interaction with a user interface to generate visual feedback on a video device.  \\
\midrule
VideoGameSeries &
  A video game series.  \\
\midrule
VisualArtwork &
  A work of art that is primarily visual in character.  \\
\midrule
Volcano &
  A volcano, like Fuji san.  \\
\midrule
Waterfall &
  A waterfall, like Niagara. \\
\midrule
WebSite &
  A WebSite is a set of related web pages and other items typically served from a single web domain and accessible via URLs.  \\
\midrule
Zoo &
  A zoo.  \\
\bottomrule
\caption{List of all semantic types in the CLSE with descriptions from \myurl{schema.org} . Entities can be associated with multiple semantic types. See~\cite{guha2016schema} or \myurl{schema.org} for more details about the semantic types.} 
\label{tab:semantic_types}
\end{longtable}
\end{small}

\clearpage
\section{Model Output Examples}
\label{app:examples}

\begin{table*}[!htb]\centering

\begin{tabular}{p{0.1\textwidth}p{0.3\textwidth}p{0.3\textwidth}p{0.2\textwidth}}
\toprule
\textbf{structured data} & \multicolumn{3}{p{0.8\textwidth}}{\texttt{service\_name: Hotels\_2 | confirm, where\_to: Amsterdam | confirm, number\_of\_adults: 5 | confirm, check\_in\_date: July 24, 2019 | confirm, check\_out\_date: April 22, 2021}} \\
\toprule
\textbf{English} & Alright. So you'd like to make a reservation for 5 in Amsterdam. You are checking in July 24, 2019 and checking out on April 22, 2021. & Alright. So you'd like to make a reservation for 5 in Amsterdam. You are checking in July 24, 2019 and checking out on April 22, 2021. & \\
\midrule
\textbf{nmt} & Bien. Donc, vous aimeriez faire une réservation pour 5 à Amsterdam. Vous vérifiez le 24 juillet 2019 et vous vérifiez le 22 avril 2021. & Good. So you would like to make a reservation for 5 in Amsterdam. You \textbf{verify} on July 24, 2019 and you \textbf{verify} on April 22, 2021. & \textit{"check in/out" mistranslated as "verify"} \\
\midrule
\textbf{d2t} & Très bien. Vous arrivez le 24 juillet 2019 et partez le 22 avril 2021. C'est bien ça? & Very well. You are arriving on July 24, 2019 and departing on April 22, 2021. \textbf{Is that correct?} & \textit{missing "for 5" and "Amsterdam", asking a question in the end} \\
\midrule
\textbf{tmpl} & D'accord. Vous souhaitez faire une réservation pour 5 personne à Amsterdam. Vous arrivez le 24 juillet 2019 et partez le 22 avril 2021. & OK. You want to make a reservation for 5 \textbf{person} in Amsterdam. You arrive on July 24, 2019 and depart on April 22, 2021. & \textit{"5 personne" is ungrammatical} \\
\midrule
\textbf{tmpl+G} & D'accord. Vous souhaitez faire une réservation pour 5 personnes à Amsterdam. Vous arrivez le 24 juillet 2019 et partez le 22 avril 2021. & OK. You want to make a reservation for 5 people in Amsterdam. You arrive on July 24, 2019 and depart on April 22, 2021. & \\
\midrule
\textbf{human} & D'accord. Vous souhaitez faire une réservation pour 5 personnes à Amsterdam. Vous arrivez le 24 juillet 2019 et partez le 22 avril 2021. & OK. You want to make a reservation for 5 people in Amsterdam. You arrive on July 24, 2019 and depart on April 22, 2021. &\\
\bottomrule
\end{tabular}
\caption{Example of different NLG system outputs in French.}
\label{tab:examples}
\end{table*}

We often see raters reporting perfect accuracy despite missed facts or new facts introduced by the model.
This is understandable, as humans are not perfect and often miss things. For example, the \textbf{d2t} output in Table~\ref{tab:examples} misses
multiple aspects, yet all three raters gave it a perfect accuracy score of $1$.
Despite high human scores, such a model would be problematic to put in front of real users, as rightfully signalled by the automatic metric(s).
Therefore, one should not completely discard automatic metrics, and use them in addition
to human scores to flag potential model issues.
Table~\ref{tab:low_bleurt_examples} exemplifies this: all results get perfect accuracy scores from the raters
whereas BLEURT successfully catches the mismatch.

\begin{table*}[h]
\begin{tabular}{p{15em}p{20em}c}\toprule
\textbf{English reference} & \textbf{d2t model prediction} & BLEURT \\
\midrule
Satire. & Je vais lancer la lecture. C'est une satire tirée de l'album de musique techno. / [I am going to start. It's a satire taken from \textbf{the techno music} album]. & 0.11 \\
\\
Please confirm the address of destination as 2901 East Hastings Street. Also for how many person do you want to book the cab and \textbf{what type of ride} you want? & Vous allez au 2901 East Hastings Street, c'est ça? Vous serez combien? / [You're going to 2901 East Hastings Street, right? How many of you?] & 0.16 \\
\\
Is your destination 2033 Camden Avenue \# F3? Please tell me the number of seats you need, as well as your ride preference. You can say Pool, Luxury or something else. & Vous souhaitez une course luxe, unisexe au 2033 Camden Avenue \# F3, pour combien de personnes? / [Would you like a luxury, \textbf{unisex} ride at 2033 Camden Avenue \#F3, for how many people?] & 0.18 \\
\bottomrule
\end{tabular}

\caption{\textbf{d2t} model outputs with the lowest BLEURT scores in French. Mismatched facts are marked in bold. All three of them get perfect accuracy scores from all three raters.
Human raters are imperfect, and for factual accuracy, there seems to be some tolerance for hallucinations or missing facts when there are many pieces of information presented in the same utterance.}
\label{tab:low_bleurt_examples}
\end{table*}

\clearpage
\twocolumn
\input{datasheet}

\end{document}

%% file: datasheet.tex
\section{Datasheet}

\href{https://arxiv.org/abs/1803.09010}{Datasheets for Datasets} %
``document [the dataset] motivation, composition, collection process, recommended uses, and so on. [They] have the potential to increase transparency and accountability within the machine learning community, mitigate unwanted biases in machine learning systems, facilitate greater reproducibility of machine learning results, and help researchers and practitioners select more appropriate datasets for their chosen tasks.''

\medskip

\definecolor{darkblue}{RGB}{46,25, 110}
 \hypersetup{
     linkcolor=red,
     filecolor=red,
     citecolor=black,      
     urlcolor=cyan,
}

\newcommand{\dssectionheader}[1]{%
   \noindent\framebox[\columnwidth]{%
      {\fontfamily{phv}\selectfont \textbf{\textcolor{darkblue}{#1}}}
   }
}

\newcommand{\dsquestion}[1]{%
    {\noindent \small \fontfamily{phv}\selectfont \textcolor{darkblue}{\textbf{#1}}}
}

\newcommand{\dsquestionex}[2]{%
    {\noindent \small \fontfamily{phv}\selectfont \textcolor{darkblue}{\textbf{#1} #2}}
}

\newcommand{\dsanswer}[1]{%
   {\noindent #1 \medskip}
}

\begin{singlespace}

\dssectionheader{Motivation}

\dsquestionex{For what purpose was the dataset created?}{Was there a specific task in mind? Was there a specific gap that needed to be filled? Please provide a description.}

\dsanswer{
CLSE was created for training, testing, and evaluating NLG systems in multiple languages,
including several low-resource ones. It allows to do sampling and slicing
by language, semantic type, or linguistic phenomena.
}

\dsquestion{Who created this dataset (e.g., which team, research group) and on behalf of which entity (e.g., company, institution, organization)?}

\dsanswer{
Google Assistant NLP Team.
}

\dsquestionex{Who funded the creation of the dataset?}{If there is an associated grant, please provide the name of the grantor and the grant name and number.}

\dsanswer{
Google Assistant NLP Team.
}

\bigskip
\dssectionheader{Composition}

\dsquestionex{What do the instances that comprise the dataset represent (e.g., documents, photos, people, countries)?}{ Are there multiple types of instances (e.g., movies, users, and ratings; people and interactions between them; nodes and edges)? Please provide a description.}

\dsanswer{
Entities of semantic types detailed in Appendix~\ref{sec:dataset_stats}.
}

\dsquestion{How many instances are there in total (of each type, if appropriate)?}

\dsanswer{
80'893 language entries (13'649 unique entities).
}

\dsquestionex{Does the dataset contain all possible instances or is it a sample (not necessarily random) of instances from a larger set?}{ If the dataset is a sample, then what is the larger set? Is the sample representative of the larger set (e.g., geographic coverage)? If so, please describe how this representativeness was validated/verified. If it is not representative of the larger set, please describe why not (e.g., to cover a more diverse range of instances, because instances were withheld or unavailable).}

\dsanswer{
The dataset represents a sample of all entities found in the Knowledge Graph. For each language and semantic type,
the sample is meant to limit
over-representation of entities with common linguistic attributes (see Figure~\ref{fig:clse_illustration}).
}

\dsquestionex{What data does each instance consist of? “Raw” data (e.g., unprocessed text or images) or features?}{In either case, please provide a description.}

\dsanswer{
See Table~\ref{tab:signature} for examples.
}

\dsquestionex{Is there a label or target associated with each instance?}{If so, please provide a description.}

\dsanswer{
No.
}

\dsquestionex{Is any information missing from individual instances?}{If so, please provide a description, explaining why this information is missing (e.g., because it was unavailable). This does not include intentionally removed information, but might include, e.g., redacted text.}

\dsanswer{
Certain linguistic attributes may not be annotated for some languages due to limited language support.
}

\dsquestionex{Are relationships between individual instances made explicit (e.g., users’ movie ratings, social network links)?}{If so, please describe how these relationships are made explicit.}

\dsanswer{
No, except for the same entity---identified by its ID---appearing for multiple languages as a separate row.
}

\dsquestionex{Are there recommended data splits (e.g., training, development/validation, testing)?}{If so, please provide a description of these splits, explaining the rationale behind them.}

\dsanswer{
No.
}

\dsquestionex{Are there any errors, sources of noise, or redundancies in the dataset?}{If so, please provide a description.}

\dsanswer{
Surface forms (entity names) and linguistic annotations were created by humans and therefore may be inaccurate
or incomplete.
}

\dsquestionex{Is the dataset self-contained, or does it link to or otherwise rely on external resources (e.g., websites, tweets, other datasets)?}{If it links to or relies on external resources, a) are there guarantees that they will exist, and remain constant, over time; b) are there official archival versions of the complete dataset (i.e., including the external resources as they existed at the time the dataset was created); c) are there any restrictions (e.g., licenses, fees) associated with any of the external resources that might apply to a future user? Please provide descriptions of all external resources and any restrictions associated with them, as well as links or other access points, as appropriate.}

\dsanswer{
The dataset is self-contained. Entity IDs refer to the Google Knowledge Graph API, but this is as an implementation detail (API stability does not affect the usefulness of the dataset).
}

\dsquestionex{Does the dataset contain data that might be considered confidential (e.g., data that is protected by legal privilege or by doctor-patient confidentiality, data that includes the content of individuals non-public communications)?}{If so, please provide a description.}

\dsanswer{
No.
}

\dsquestionex{Does the dataset contain data that, if viewed directly, might be offensive, insulting, threatening, or might otherwise cause anxiety?}{If so, please describe why.}

\dsanswer{
No, to the best of our knowledge.
}

\dsquestionex{Does the dataset relate to people?}{If not, you may skip the remaining questions in this section.}

\dsanswer{
Some entities in the dataset are of semantic type ``Person.''
}

\dsquestionex{Does the dataset identify any subpopulations (e.g., by age, gender)?}{If so, please describe how these subpopulations are identified and provide a description of their respective distributions within the dataset.}

\dsanswer{
No.
}

\dsquestionex{Is it possible to identify individuals (i.e., one or more natural persons), either directly or indirectly (i.e., in combination with other data) from the dataset?}{If so, please describe how.}

\dsanswer{
Yes, people with a Knowledge Graph entry can be uniquely identified.
}

\dsquestionex{Does the dataset contain data that might be considered sensitive in any way (e.g., data that reveals racial or ethnic origins, sexual orientations, religious beliefs, political opinions or union memberships, or locations; financial or health data; biometric or genetic data; forms of government identification, such as social security numbers; criminal history)?}{If so, please provide a description.}

\dsanswer{
No, to the best of our knowledge.
}

\bigskip
\dssectionheader{Collection Process}

\dsquestionex{How was the data associated with each instance acquired?}{Was the data directly observable (e.g., raw text, movie ratings), reported by subjects (e.g., survey responses), or indirectly inferred/derived from other data (e.g., part-of-speech tags, model-based guesses for age or language)? If data was reported by subjects or indirectly inferred/derived from other data, was the data validated/verified? If so, please describe how.}

\dsanswer{
The data was curated by linguists. See Section~\ref{sec:clse} for more details.
}

\dsquestionex{What mechanisms or procedures were used to collect the data (e.g., hardware apparatus or sensor, manual human curation, software program, software API)?}{How were these mechanisms or procedures validated?}

\dsanswer{
The data was curated in spreadsheets and text files and, as a rule, reviewed by another linguist.
}

\dsquestion{If the dataset is a sample from a larger set, what was the sampling strategy (e.g., deterministic, probabilistic with specific sampling probabilities)?}

\dsanswer{
Exact details of the sampling procedure cannot be disclosed at the moment to preserve anonymity and to comply
with internal policies of the authors' organizations.
}

\dsquestion{Who was involved in the data collection process (e.g., students, crowdworkers, contractors) and how were they compensated (e.g., how much were crowdworkers paid)?}

\dsanswer{
Contractors. Each contract is reviewed, approved, and executed according to the strict company policies.
}

\dsquestionex{Over what timeframe was the data collected? Does this timeframe match the creation timeframe of the data associated with the instances (e.g., recent crawl of old news articles)?}{If not, please describe the timeframe in which the data associated with the instances was created.}

\dsanswer{
The bulk of linguistic data was collected over the years 2020 and 2021. Semantic type associations were retrieved from the Google Knowledge Graph API on 2022-05-18.
}

\dsquestionex{Were any ethical review processes conducted (e.g., by an institutional review board)?}{If so, please provide a description of these review processes, including the outcomes, as well as a link or other access point to any supporting documentation.}

\dsanswer{
Yes. The dataset description was improved and this datasheet was created as an outcome.
}

\dsquestionex{Does the dataset relate to people?}{If not, you may skip the remaining questions in this section.}

\dsanswer{
Some entities in the dataset are of semantic type ``Person.'' These are limited to individuals (alive,
dead, or fictional) who are popular enough to have a Knowledge Graph entry.
}

\dsquestion{Did you collect the data from the individuals in question directly, or obtain it via third parties or other sources (e.g., websites)?}

\dsanswer{
No.
}

\dsquestionex{Were the individuals in question notified about the data collection?}{If so, please describe (or show with screenshots or other information) how notice was provided, and provide a link or other access point to, or otherwise reproduce, the exact language of the notification itself.}

\dsanswer{
No.
}

\dsquestionex{Did the individuals in question consent to the collection and use of their data?}{If so, please describe (or show with screenshots or other information) how consent was requested and provided, and provide a link or other access point to, or otherwise reproduce, the exact language to which the individuals consented.}

\dsanswer{
N/A.
}

\dsquestionex{If consent was obtained, were the consenting individuals provided with a mechanism to revoke their consent in the future or for certain uses?}{If so, please provide a description, as well as a link or other access point to the mechanism (if appropriate).}

\dsanswer{
N/A.
}

\dsquestionex{Has an analysis of the potential impact of the dataset and its use on data subjects (e.g., a data protection impact analysis) been conducted?}{If so, please provide a description of this analysis, including the outcomes, as well as a link or other access point to any supporting documentation.}

\dsanswer{
N/A.
}

\bigskip
\dssectionheader{Preprocessing/cleaning/labeling}

\dsquestionex{Was any preprocessing/cleaning/labeling of the data done (e.g., discretization or bucketing, tokenization, part-of-speech tagging, SIFT feature extraction, removal of instances, processing of missing values)?}{If so, please provide a description. If not, you may skip the remainder of the questions in this section.}

\dsanswer{
No.
}

\dsquestionex{Was the “raw” data saved in addition to the preprocessed/cleaned/labeled data (e.g., to support unanticipated future uses)?}{If so, please provide a link or other access point to the “raw” data.}

\dsanswer{
N/A.
}

\dsquestionex{Is the software used to preprocess/clean/label the instances available?}{If so, please provide a link or other access point.}

\dsanswer{
N/A.
}

\bigskip
\dssectionheader{Uses}

\dsquestionex{Has the dataset been used for any tasks already?}{If so, please provide a description.}

\dsanswer{
Yes, for experiments in Section~\ref{sec:experiment}.
}

\dsquestionex{Is there a repository that links to any or all papers or systems that use the dataset?}{If so, please provide a link or other access point.}

\dsanswer{
No.
}

\dsquestion{What (other) tasks could the dataset be used for?}

\dsanswer{
E.g., for balancing machine translation data.
}

\dsquestionex{Is there anything about the composition of the dataset or the way it was collected and preprocessed/cleaned/labeled that might impact future uses?}{For example, is there anything that a future user might need to know to avoid uses that could result in unfair treatment of individuals or groups (e.g., stereotyping, quality of service issues) or other undesirable harms (e.g., financial harms, legal risks) If so, please provide a description. Is there anything a future user could do to mitigate these undesirable harms?}

\dsanswer{
The linguistic attributes are provided ``as is'' and may be innacurate or incomplete.
}

\dsquestionex{Are there tasks for which the dataset should not be used?}{If so, please provide a description.}

\dsanswer{
The dataset should not be used to infer non-linguistic properties of entities. In particular,
the linguistic attributes are not appropriate proxy data to infer a person's aliveness or gender.
}

\bigskip
\dssectionheader{Distribution}

\dsquestionex{Will the dataset be distributed to third parties outside of the entity (e.g., company, institution, organization) on behalf of which the dataset was created?}{If so, please provide a description.}

\dsanswer{
Yes.
}

\dsquestionex{How will the dataset will be distributed (e.g., tarball on website, API, GitHub)}{Does the dataset have a digital object identifier (DOI)?}

\dsanswer{
As a CSV file retrievable from \myurl{github.com/google-research-datasets/clse}.
}

\dsquestion{When will the dataset be distributed?}

\dsanswer{
Upon acceptance of the publication.
}

\dsquestionex{Will the dataset be distributed under a copyright or other intellectual property (IP) license, and/or under applicable terms of use (ToU)?}{If so, please describe this license and/or ToU, and provide a link or other access point to, or otherwise reproduce, any relevant licensing terms or ToU, as well as any fees associated with these restrictions.}

\dsanswer{
Yes, \href{https://creativecommons.org/licenses/by/4.0/}{CC-BY} license.
}

\dsquestionex{Have any third parties imposed IP-based or other restrictions on the data associated with the instances?}{If so, please describe these restrictions, and provide a link or other access point to, or otherwise reproduce, any relevant licensing terms, as well as any fees associated with these restrictions.}

\dsanswer{
No, to the best of our knowledge.
}

\dsquestionex{Do any export controls or other regulatory restrictions apply to the dataset or to individual instances?}{If so, please describe these restrictions, and provide a link or other access point to, or otherwise reproduce, any supporting documentation.}

\dsanswer{
No, to the best of our knowledge.
}

\bigskip
\dssectionheader{Maintenance}

\dsquestion{Who will be supporting/hosting/maintaining the dataset?}

\dsanswer{
The authors of this publication.
}

\dsquestion{How can the owner/curator/manager of the dataset be contacted (e.g., email address)?}

\dsanswer{
Yes, by email or any other contact point provided at \myurl{github.com/google-research-datasets/clse}.
}

\dsquestionex{Is there an erratum?}{If so, please provide a link or other access point.}

\dsanswer{
No.
}

\dsquestionex{Will the dataset be updated (e.g., to correct labeling errors, add new instances, delete instances)?}{If so, please describe how often, by whom, and how updates will be communicated to users (e.g., mailing list, GitHub)?}

\dsanswer{
No updates are planned at the moment. If any is made, it will be communicated at \myurl{github.com/google-research-datasets/clse}.
}

\dsquestionex{If the dataset relates to people, are there applicable limits on the retention of the data associated with the instances (e.g., were individuals in question told that their data would be retained for a fixed period of time and then deleted)?}{If so, please describe these limits and explain how they will be enforced.}

\dsanswer{
N/A.
}

\dsquestionex{Will older versions of the dataset continue to be supported/hosted/maintained?}{If so, please describe how. If not, please describe how its obsolescence will be communicated to users.}

\dsanswer{
Yes.
}

\dsquestionex{If others want to extend/augment/build on/contribute to the dataset, is there a mechanism for them to do so?}{If so, please provide a description. Will these contributions be validated/verified? If so, please describe how. If not, why not? Is there a process for communicating/distributing these contributions to other users? If so, please provide a description.}

\dsanswer{
Please, contact the dataset mainteners using the contact information above.
}

\end{singlespace}